# Semi-Optimal Edge Detector based on Simple Standard Deviation with Adjusted Thresholding

Firas A. Jassim
Management Information Systems Department,
Irbid National University,
Irbid 2600, Jordan.

## ABSTRACT

This paper proposes a novel method which combines both median filter and simple standard deviation to accomplish an excellent edge detector for image processing. First of all, a denoising process must be applied on the grey scale image using median filter to identify pixels which are likely to be contaminated by noise. The benefit of this step is to smooth the image and get rid of the noisy pixels. After that, the simple statistical standard deviation could be computed for each 2×2 window size. If the value of the standard deviation inside the 2×2 window size is greater than a predefined threshold, then the upper left pixel in the 2×2 window represents an edge. The visual differences between the proposed edge detector and the standard known edge detectors have been shown to support the contribution in this paper.

## General Terms

Image Processing and Pattern Recognition.

## Keywords

Computer vision, edge detection, median filter, standard deviation.

## 1. INTRODUCTION

Edge detection is one of the most important techniques that have been commonly implemented in image processing and computer vision. There are many image processing applications that are based on edge detection like image segmentation, registration and identification. The concept of the edge in an image is the most fundamental feature of the image because the edge contains valuable information about the internal objects inside image. Hence, edge detection is one of the key research works in image processing [6]. Detection of edges in an image is a very important step towards understanding image features. The process of recognition of objects in an image is referred to as image understanding system [7]. Therefore, other image processing applications such as segmentation, identification, and object recognition can take place whenever edges of an object are detected [8]. A huge number of edge detectors have been developed from different perspectives [11]. The main question here is: which edge detector can engender better edge detection results?. In the literature, there are some techniques developed to achieve this task such as Sobel, Prewitt, Laplacian, Laplacian of Gaussian (LOG), and Canny which is used to be the optimal edge detector [3].

Recently, many researchers have developed novel edge detectors that reflect the flexibility of edge detection method by the accommodation with other fields of science. The integration of fuzzy theory with edge detection has been discussed by [1]. Also, the embedding of artificial neural networks was researched by [17]. Furthermore, other algorithms for edge detection could be found in [4][17][16]. A good Comparison of various edge detectors may be found in [12][18]. Edge detection is used for object detection which serves various applications in image processing [10].

The organization of this paper is as follows: In section 2 a preliminaries about edge detection method have been discussed. Also, in section 3, the main contribution of this paper has been presented and that is using simple standard deviation for edge detection. In section 4, medina filter was implemented into the proposed edge detector. The experimental results with ocular examples have been showed in section 5. Finally, the main conclusions about the proposed algorithm were demonstrated in section 6.

## 2. EDGE DETECTION PRELIMINARIES

Edges consist of meaningful features and contained significant information. Applying an edge detector to an image may significantly reduce the amount of data to be processed and may therefore filter out information that may be regarded as less relevant, while preserving the important structural properties of an image [18]. The essential notion of the majority edge detectors is to determine some boundary information in an image that represents the image's interior objects. According to [8], edge is a set on connected pixels that lie on the boundary between two regions. Also, an edge in an image is a contour across which the brightness of the image changes suddenly in amount [13]. Edge refers to the pixel set whose gray level or gradient direction occurs sudden change and usually evinces linear feature [9]. Generally, an edge is defined as the borderline pixels that connect two mutually exclusive regions which differ in their luminance and tristimulus values [15]. The edge of an object is reflected in the discontinuity of the gray [6]. Hence, the fundamental method of edge detection is the local operator edge detection method. In this method, pixel in a region must be compared with its neighbors for the differences in order to detect the edge [6]. The detection operation starts with the inspection of the local discontinuity at each pixel in the region. Consequently, the determination of an edge is based on some characteristics that are amplitude, location and orientation of a region [5]. Therefore, based on these characteristics, the investigator has to examine each pixel to determine whether it is an edge or not [15].

Usually, the main principle for edge detection in the literature is based on the gradient vector for an image. Hence, the traditional technique for edge detection may be lying in the filed of the calculus of variation. In this article, an absolutely different approach for edge detection had been established which was based on statistics instead of the traditional calculus of variation. Therefore, the contribution in this paper may be treated as statistical edge detector.

## 3. EDGE DETECTION BASED ON STANDARD DEVIATION

The Standard Deviation is a measure of how spreads out numbers are. Also, it is the measure of the dispersion of a set of data from its mean. The more spread apart the data, the higher the deviation. A window of size 2×2 will be implemented to detect the candidate point (i,j) as an edges, figure (1).

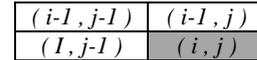

**Fig 1. Candidate edge (i,j) in a 2×2 window size**

According to figure (2), six types of edges were used which are the diagonal edges (2.a), (2.b), (2.c), and (2.d) and the horizontal edge (2.e) and the vertical edge (2.f). On the other hand, the 2×2 window does not contain any edge if it is looks like figure (2.g). In fact, the value of the standard deviation in the 2×2 window will vary according to the dispersion between its pixels.

Now, as an illustrative example will be discussed in details to support the proposed technique. Hence, a random 10×10 window from Lena image was taken as shown in figure (3).

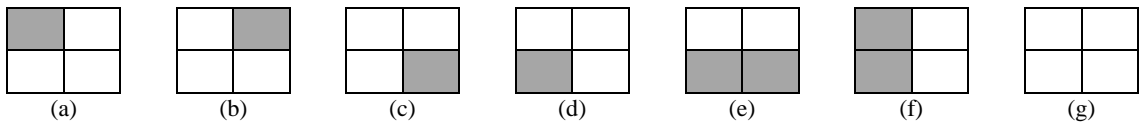

**Fig 2. (a), (b), (c), and (d) are diagonal edges (e) horizontal edge (f) vertical edge (g) no edge**

| 201 | 205 | 182 | 134 | 94  | 94  | 115 | 120 | 116 | 111 |
|-----|-----|-----|-----|-----|-----|-----|-----|-----|-----|
| 204 | 172 | 113 | 83  | 93  | 103 | 96  | 105 | 104 | 102 |
| 159 | 103 | 80  | 86  | 97  | 100 | 100 | 95  | 101 | 103 |
| 114 | 83  | 76  | 84  | 88  | 83  | 78  | 71  | 77  | 81  |
| 79  | 72  | 75  | 81  | 80  | 72  | 65  | 52  | 56  | 59  |
| 71  | 71  | 72  | 72  | 68  | 65  | 63  | 51  | 51  | 52  |
| 68  | 69  | 64  | 58  | 54  | 54  | 55  | 56  | 54  | 52  |
| 66  | 67  | 60  | 52  | 49  | 48  | 48  | 53  | 52  | 51  |
| 67  | 64  | 55  | 49  | 50  | 50  | 48  | 49  | 49  | 50  |
| 69  | 59  | 46  | 41  | 47  | 51  | 50  | 48  | 50  | 51  |

**Fig 3. A random 10×10 window from Lena image**

Clearly, this block contains edges because of the high variation between its pixels. Therefore, a 2×2 window size was implemented, here, arbitrary for the matter of shortening. The standard deviation for the upper left 2×2 window was found to be 15.7586 which seemly to be some what high value. Subsequently, the other values were measured (according to the upper left pixel in each 2×2 window) and presented in table (1).

**Table 1. Standard deviation values for arbitrary 2×2 window sizes**

| Upper left point | Standard deviation |
|------------------|--------------------|
| 201              | 15.7586            |
| 172              | 39.1833            |
| 134              | 22.5536            |
| 103              | 2.8723             |
| 115              | 10.6771            |
| 116              | 6.4485             |
| 101              | 13.4040            |
| 83               | 4.6547             |
| 67               | 5.1962             |
| 49               | 0.8165             |

As a result, a threshold may be applied to determine weather a 2×2 window contains an edge or not. The smaller the value of the standard deviation yields there is no edge inside the 2×2 window. Alternatively, a higher value of the standard deviation implies that there is a definite edge. Mathematically speaking, using the concept of try and error, the value of the suitable threshold that can be used as a general threshold for almost type of digital image was found to be near 7, i.e. from 4 to 9 will gives better results.

## 4. Median Filter

Median filter is widely used in impulse noise removal methods due to its denoising ability and computational efficiency [8]. Therefore, applying of traditional median filter for removal of such type of noise gives relatively acceptable results, which are shown in restoring of brightness drops, objects edges and local peaks in noise corrupted images [2]. Hence, the edge detection process may be preceded by some denoising process to smooth the image. These two processes should be compatible with each other to produce good results. According to figure (4), it is clear that the proposed edge detector gives better results than Sobel and Canny edge detectors when applying 10% Salt and Peppers noise. Obviously, the visual differences between figures (4.e) and (4.f) are distinct. Since the proposed edge detector is based on the concept of the deviation between 2×2 window pixels, then the standard deviation within the 2×2 window was calculated before removing the Salt and Peppers noise. This is a critical situation because the noisy pixels were treated as noise-free pixels when calculating the standard deviation. On the other hand, figure (4.f), the proposed edge detector was implemented by, firstly, removing the noisy pixels to get rid of the high dispersion between 2×2 window pixels. This step would highly enhance the proposed edge detector by preserving the edges as possible. As a result, the compatibility between the median filter and the proposed edge detector is needed to accomplish the proposed edge detection process.

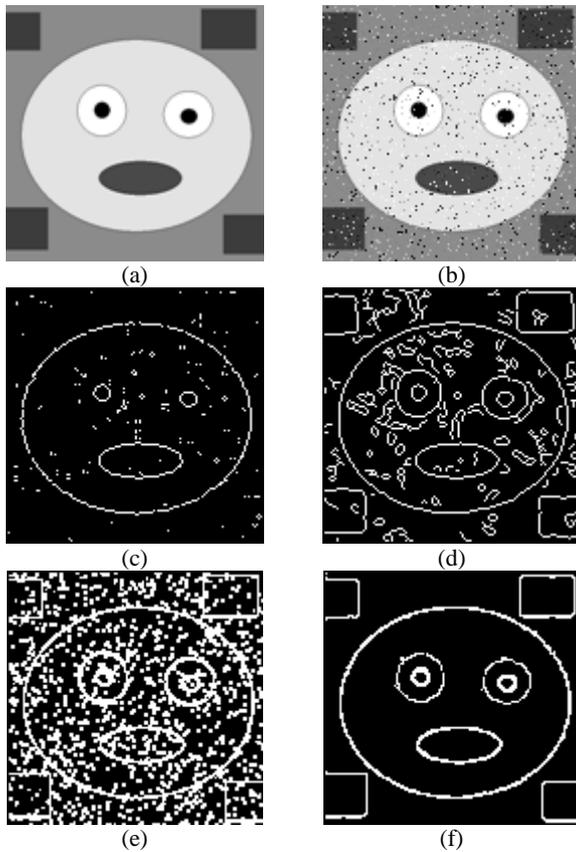

**Fig 4. (a) Original 128×128 bmp image (b) 10% Salt and Peppers noise (c) Sobel (d) Canny (e) Proposed without median filter (f) Proposed with median filter**

## 5. EXPERIMENTAL RESULTS

The proposed algorithm was simulated in MATLAB and applied on two categories of images. The first category is the natural images (fig. 5) and the second one is the text images (fig. 6). A collection of gray level test images with different sizes and resolutions were detected. The results were compared with the most common edge detection methods such as Sobel [14] and the optimal Canny's edge detector [3].

According to figure (5), it is clear that the proposed edge detector had demonstrated the use of standard deviation for edge detection. The proposed edge detector is near or semi-optimal edge detector (Canny). The use of the word semi here was because of the thick edges produced by the proposed edge detector which is the only disadvantage. On the other hand, the proposed edge detector yields better results than the traditional edge detector (Sobel). Consequently, the proposed edge detector is something between the traditional and the optimal edge detectors. Hence, the word semi has been appeared to accommodate the situation correctly.

According to figures (5) and (6), the proposed edge detector produces explicit results than the optimal Canny's edge detector and the traditional Sobel's edge detector.

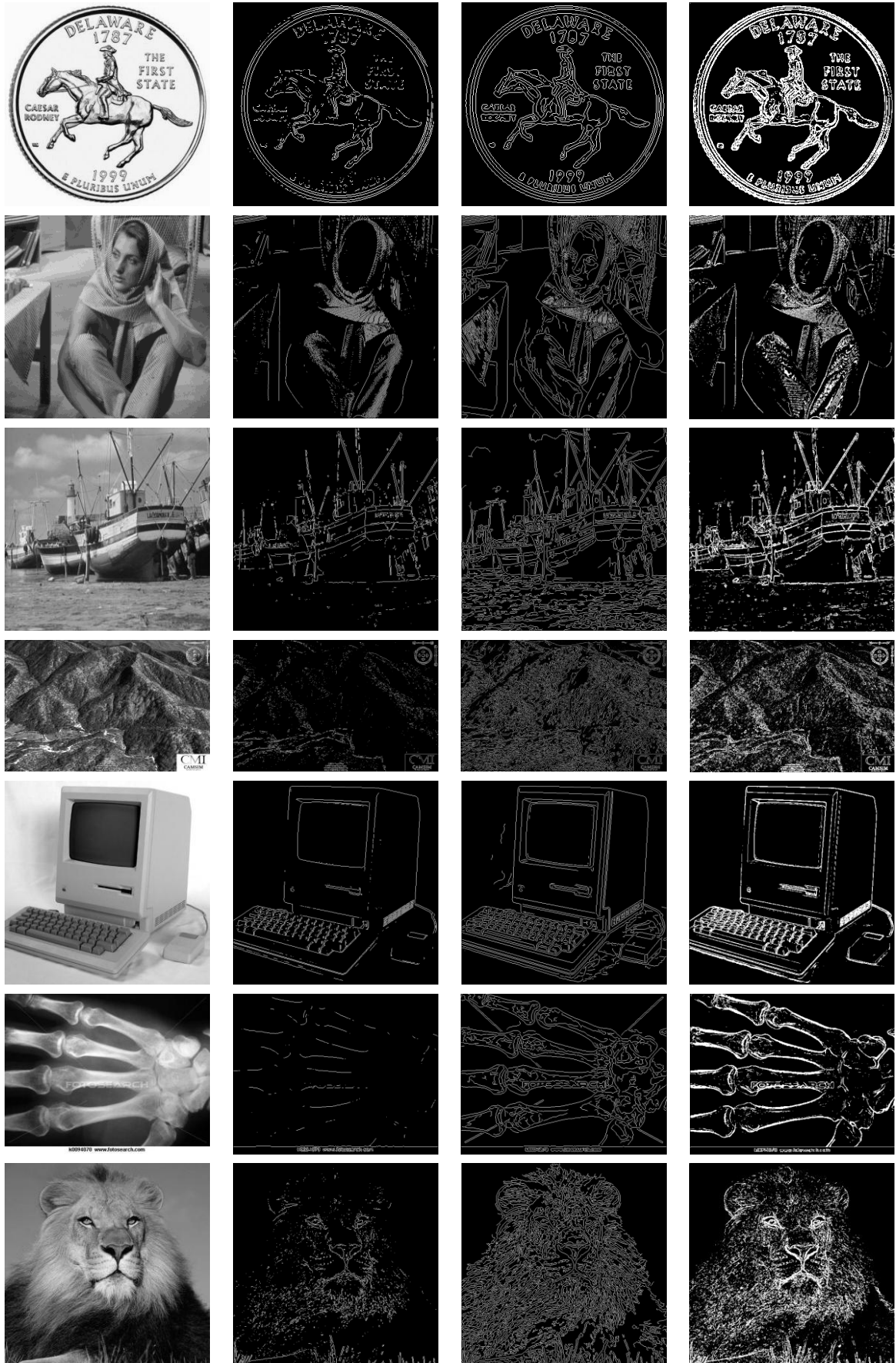

**Fig 5. (Columns from left to right) original image, Sobel, Canny, and Proposed (Threshold =7)**

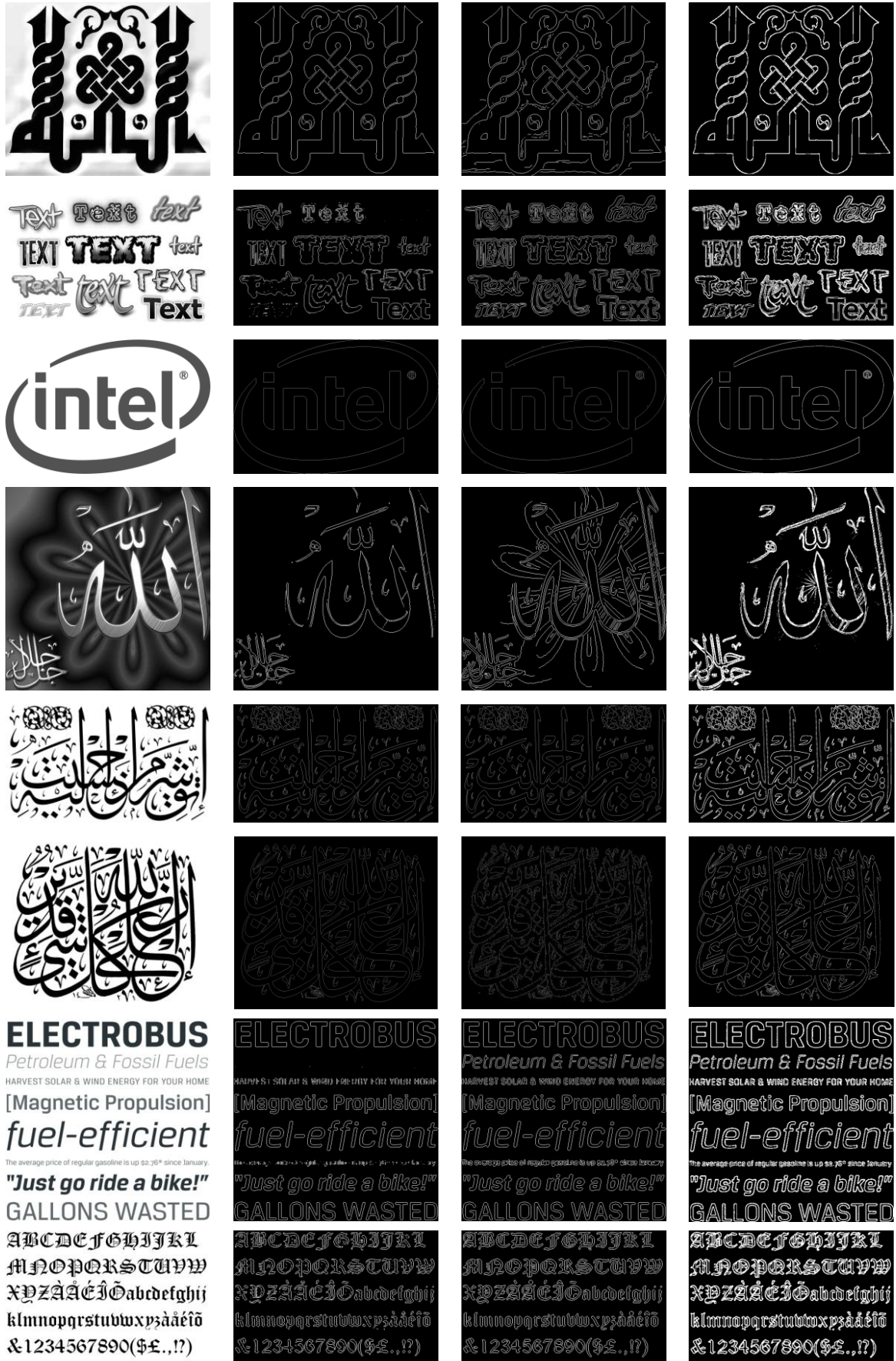

**Fig 6. (Columns from left to right) original image, Sobel, Canny, and Proposed (Threshold =7)**

## 6. CONCLUSIONS

In this paper, a new methodology and a framework for evaluating edge detection by simple statistical standard deviation had been presented. The proposed method was tested on binary level test images and compared with two classical edge detectors which are Sobel and Canny. Experiment results have demonstrated that the proposed algorithm for edge detection produces satisfactory results for different gray level digital images. Furthermore, the proposed edge detector gives better results on text images compared with traditional edge detectors rather than natural images. Therefore, a good recommendation is that the proposed edge detector is suitable for text images that contain any kind of text. Another benefit for the proposed edge detector is the easy implementation.

## 7. ACKNOWLEDGMENTS

The author would like to thank Mrs. Hind E. Qassim for her excellent notes and support for preparing this paper.

## 8. REFERENCES

[1] Becerikli Y. and Karan T. M. 2005. A New Fuzzy Approach for Edge Detection. IWANN 2005, LNCS 3512, 943–951.

[2] Bovik A. C., Huang T. S., and Munson D. C. 1985. Edge-sensitive image restoration using order constrained least squares methods. IEEE Trans. Acoust., Speech, Signal Processing, vol. 33, pp. 1253–1263.

[3] Canny J 1986. A computational approach to edge detection. IEEE Transactions on Pattern Analysis and Machine Intelligence, 8(6):679–698.

[4] Desolneux A., Moisan L. And Morel J.-M. 2001. Edge Detection by Helmholtz Principle, Journal of Mathematical Imaging and Vision 14: 271–284.

[5] Frei W. and Chen C. 1977. Fast Boundary Detection: A Generalization and New Algorithm. IEEE Trans. Computers, vol. C-26, no. 10, 988-998.

[6] Gao W., Yang L., Zhang X. and Liu H. 2010. An Improved Sobel Edge Detection. IEEE international conference on computer science and information technology (ICCSIT), vol. 5, 67-71.

[7] Giannarou S. and Stathaki T. 2011. Optimal edge detection using multiple operators for image understanding. EURASIP Journal on Advances in Signal Processing 2011, 28.

[8] Gonzalez R. C. and Woods R. E. 2001. Digital Image Processing. Upper Saddle River, NJ: Prentice-Hall.

[9] Jing L., Peikang H., Xiaohu W. and Xudong P. 2009. Image edge detection based on beamlet transform, Journal of Systems Engineering and Electronics, Vol. 20, No. 1, 1–5.

[10] Lei Z., Shouping D. and Honglian M., 2008. Recent Methods and Applications on Image Edge Detection. Proceedings of the 2008 International Workshop on Education Technology and Training & 2008 International Workshop on Geoscience and Remote Sensing, vol. 1, 332-335.

[11] Maini R. and Aggarwal H, 2009. Study and Comparison of Various Image Edge Detection Techniques. International Journal of Image Processing (IJIP), vol. 3, issue 1, 1-11.

[12] Nadernejad E. 2008. Edge Detection Techniques: Evaluations and Comparisons. Applied Mathematical Sciences, Vol. 2, no. 31, 1507 – 1520.

[13] Ritter G. X. and Joseph N. Wilson 2000. Handbook of Computer Vision Algorithms in Image Algebra, CRC Press.

[14] Sobel I. E. 1970. Camera models and machine perception. Ph.D. dissertation, Stanford University, Stanford, Calif, USA.

[15] Tzu-Heng Henry Lee. Edge Detection Analysis. Graduate Institute of Communication Engineering, National Taiwan University, Taipei, Taiwan, ROC.

[16] Yi S., Labate D., Easley G. R. and Krim H. 2009. A Shearlet Approach to Edge Analysis and Detection. IEEE Trans. Image Proc. 18(5) 929–941.

[17] Zheng L. He X. Edge, 2004. Detection Based on Modified BP Algorithm of ANN. Proceeding VIP 2003 of the Pan-Sydney area workshop on visual information processing, 119-122.

[18] Ziou D. and Tabbone S. 1998. Edge Detection Techniques - An Overview. International Journal of Pattern Recognition and Image Analysis, vol. 8,537-559.

## AUTHORS' PROFILE

**Firas A. Jassim** received the BS degree in mathematics and computer applications from Al-Nahrain University, Baghdad, Iraq in 1997, and the MS degree in mathematics and computer applications from Al-Nahrain University, Baghdad, Iraq in 1999 and the PhD degree in computer information systems from the university of banking and financial sciences, Amman, Jordan in 2012. Now, he is working as an assistant professor with Management Information System Department at Irbid National University, Irbid, Jordan. His research interests are Image processing, image compression, image enhancement, image interpolation and simulation.